\documentclass{egpubl}
\usepackage{EGUKCGVC2024}
 
\ConferenceSubmission   
\WsPaper           

\usepackage[T1]{fontenc}
\usepackage{dfadobe}  

\usepackage{amsmath}
\usepackage{amssymb}
\usepackage{lipsum}
\usepackage{float}

\biberVersion
\BibtexOrBiblatex
\usepackage[backend=biber,bibstyle=EG,citestyle=alphabetic,backref=true]{biblatex} 
\electronicVersion
\PrintedOrElectronic

\ifpdf \usepackage[pdftex]{graphicx} \pdfcompresslevel=9
\else \usepackage[dvips]{graphicx} \fi

\usepackage{egweblnk} 

\title[]%
      {The Misclassification Likelihood Matrix: \\ Some Classes Are More Likely To Be Misclassified Than Others}

\author[D. Sikar, A. d'Avila Garcez, R. Bloomfield, T. Weyde, K. Peeroo, N. Singh, M. Hutchinson, D. Laksono,  M. Reljan-Delaney]
{\parbox{\textwidth}{\centering D. Sikar, A. d'Avila Garcez\orcid{0000-0001-7375-9518}, R. Bloomfield, T. Weyde\orcid{0000-0001-8028-9905}, K. Peeroo, N. Singh, M. Hutchinson, D. Laksono,  M. Reljan-Delaney
        }
        \\
{\parbox{\textwidth}{\centering Department of Computer Science, City St George's, University of London
       }
}
}

\begin{document}


\maketitle

\begin{abstract}
    This study introduces the Misclassification Likelihood Matrix (MLM) as a novel tool for quantifying the reliability of neural network predictions under distribution shifts. The MLM is obtained by leveraging softmax outputs and clustering techniques to measure the distances between the predictions of a trained neural network and class centroids. By analyzing these distances, the MLM provides a comprehensive view of the model's misclassification tendencies, enabling decision-makers to identify the most common and critical sources of errors. The MLM allows for the prioritization of model improvements and the establishment of decision thresholds based on acceptable risk levels. The approach is evaluated on the MNIST dataset using a Convolutional Neural Network (CNN) and a perturbed version of the dataset to simulate distribution shifts. The results demonstrate the effectiveness of the MLM in assessing the reliability of predictions and highlight its potential in enhancing the interpretability and risk mitigation capabilities of neural networks. The implications of this work extend beyond image classification, with ongoing applications in autonomous systems, such as self-driving cars, to improve the safety and reliability of decision-making in complex, real-world environments.


\end{abstract}      

\section{Introduction}

In a multiclass classification setting, the severity and consequences of misclassifications can vary significantly depending on the specific problem domain and the nature of the classes involved. Some misclassifications may lead to more serious repercussions than others, making it crucial to consider the relative importance of different types of errors.

For example, in medical diagnosis, misclassifying a malignant tumor as benign can have life-threatening consequences, as it may delay necessary treatment. On the other hand, misclassifying a benign tumor as malignant may lead to unnecessary interventions and patient anxiety, but the consequences are generally less severe compared to a false negative. In this context, false negatives are considered more hazardous than false positives.

Similarly, in autonomous driving systems, misclassifying a pedestrian as a non-pedestrian object can result in a collision and potential loss of life, while misclassifying a non-pedestrian object as a pedestrian may cause the vehicle to unnecessarily brake or take evasive action, which is less dangerous.

Additionally, financial systems, where errors might result in significant economic losses; or legal applications, where misjudgments could affect individuals' lives.

Researchers have studied and addressed the varying hazards of misclassifications in different domains. \cite{kuncheva2006measures} explore the relationship between diversity measures and ensemble accuracy in classifier ensembles. They discuss the importance of considering the costs of different types of errors and how they can be incorporated into the design and evaluation of classifier ensembles.
\cite{mozannar2020consistent} propose a framework for consistent cost-sensitive learning with nonlinear loss functions. They address the issue of varying misclassification costs and present algorithms that can effectively handle such scenarios, enabling the learning of classifiers that are sensitive to the relative hazards of different types of errors.

Approaches such as Multi-class Difference in Confidence and Accuracy (MDCA) \cite{hebbalaguppe2022stitch}, aim to align CNN-like models' predicted confidence scores with their actual accuracy across all classes, not just the predicted class (DCA).

Ensuring the reliability of automated decision-making systems is crucial in high-stakes domains where errors can lead to severe consequences \cite{amodei2016concrete}. Machine learning models, despite their impressive performance, are susceptible to making inaccurate predictions when presented with data that deviates from the training distribution, a problem known as distribution shift \cite{hendrycks2021many, quinonero2009dataset}.

Distribution shift is a critical challenge in machine learning that occurs when the statistical properties of the data used to train a model differ from those encountered during deployment or testing \cite{quinonero2009dataset, hendrycks2019benchmarking}. This mismatch can lead to a significant decline in model performance, as the model's learned patterns may not generalize well to the new data distribution.

Distribution shift can manifest in various forms. Covariate shift occurs when the distribution of input features changes between training and deployment, while the relationship between features and labels remains the same \cite{shimodaira2000improving}. Concept drift refers to situations where the relationship between input features and the target variable changes over time, even if the input distribution remains constant \cite{gama2014survey}. Domain shift encompasses changes in both the input distribution and the relationship between inputs and outputs, often seen when applying a model to a new domain \cite{patel2015visual}. These different types of shifts present unique challenges and require tailored approaches to address them effectively.

To quantify and mitigate distribution shift, researchers have developed a range of sophisticated techniques. Statistical divergence measures, such as Kullback-Leibler divergence \cite{kullback1951information} or Maximum Mean Discrepancy \cite{gretton2012kernel}, help quantify the difference between two probability distributions, allowing researchers to detect and measure distribution shift. These measures provide valuable insights into the extent of the shift and can guide mitigation strategies.

Domain adaptation techniques aim to adapt a model trained on one domain (source) to perform well on a different but related domain (target) \cite{wang2018deep, sugiyama2007covariate, pan2009survey, ganin2016domain}. These methods include transfer learning, where knowledge gained from one task is applied to a different but related task, adversarial training, which aims to learn domain-invariant features, and importance weighting, which adjusts the contribution of training samples based on their relevance to the target domain.

Adaptive algorithms are designed to continuously update and adapt to changing data distributions \cite{lu2018learning, baena2006early}. These approaches often involve online learning or incremental learning techniques that can adjust model parameters as new data becomes available. Such algorithms are particularly useful in dynamic environments where the data distribution may evolve over time.

Quantifying the likelihood of misclassifications under distribution shift is crucial for assessing and mitigating risks in real-world deployments. By estimating the probability of incorrect predictions, decision-makers can make informed choices about when to rely on automated systems and when to defer to human judgment. This is particularly important in dynamic environments where the data distribution may evolve over time, necessitating continuous monitoring and adaptation of the models.
    

\section{Context}


\subsection{Softmax Output And Environmental Aspects}

The softmax output of a classifier represents the predicted probabilities for each class. In an $n$-class classification problem, the softmax function takes the logits (raw outputs) of the classifier and transforms them into a probability distribution over the $n$ classes \cite{goodfellow2016deep}. The softmax output is a vector of length $n$, where each element represents the predicted probability of the input belonging to the corresponding class. The probabilities in the softmax output sum up to 1, indicating the relative confidence of the classifier in each class prediction. The class with the highest probability is typically considered the predicted class. The softmax output provides a more informative and interpretable representation of the classifier's predictions compared to the raw logits.

When making a final decision, it is common practice to select only the class with the highest probability, effectively discarding the remaining $(n-1)/n$ of the softmax output \cite{gal2016uncertainty}. This approach, while straightforward and widely used, raises concerns about the inefficient use of the model's output and its potential environmental implications.

By considering only the class prediction, we are essentially discarding a significant portion of the information generated by the model. In a scenario with a large number of classes, this can lead to a substantial waste of computational resources and energy \cite{strubell2019energy}. The discarded softmax probabilities, which contain valuable information about the model's uncertainty and the relationships between classes, are effectively consigned to a digital landfill, contributing to the growing problem of electronic waste and energy consumption in the field of machine learning \cite{schwartz2020green}.

The environmental impact of this practice becomes more pronounced when considering the increasing scale and complexity of modern machine learning models. As the number of classes grows and models become more sophisticated, the amount of discarded information also increases, leading to a larger digital carbon footprint \cite{lacoste2019quantifying}. This is particularly concerning given the rapid growth of machine learning applications in various domains, from image and speech recognition to natural language processing and autonomous systems \cite{thompson2020computational}.

Certain approaches lend themselves well to make better use of the entire softmax output. One such approach is to incorporate uncertainty estimation techniques, such as Monte Carlo dropout \cite{gal2016dropout} or ensemble methods \cite{lakshminarayanan2017simple}, which leverage the full probability distribution to quantify the model's confidence in its predictions. By considering the uncertainty information, systems can make more informed decisions and avoid discarding potentially useful information \cite{kendall2017uncertainties}.

Another approach is to develop more efficient and environmentally-friendly machine learning techniques, such as model compression \cite{han2015deep}, quantization \cite{gholami2021survey}, and energy-aware training \cite{garcia2021estimation}. These methods aim to reduce the computational and memory requirements of models while maintaining their performance, thereby reducing the environmental impact of machine learning systems.

Furthermore, there is a growing awareness of the need for sustainable AI practices and the development of green AI frameworks \cite{schwartz2020green}. These initiatives focus on designing and implementing machine learning systems that prioritize energy efficiency, resource conservation, and environmental sustainability. Sustainable AI practices can help mitigate the environmental impact of discarded softmax outputs and other inefficiencies in machine learning pipelines.


\subsection{Using All Softmax Output Predictions}

Utilizing the entire set of softmax prediction probabilities, instead of solely depending on the maximum output, has been widely investigated to improve the safety, robustness, and trustworthiness of machine learning models. By considering the complete distribution of class predictions, more reliable and informative prediction pipelines can be developed, enabling techniques such as uncertainty quantification, anomaly detection, and robustness against adversarial attacks \cite{kendall2017uncertainties, lakshminarayanan2017simple, hendrycks17baseline, goodfellow2014explaining, szegedy2013intriguing}.

The softmax probabilities provide valuable information for interpretability and explainability of the model's decisions, facilitating human-machine collaboration and trust \cite{ribeiro2016should, doshi2017towards}. The importance of leveraging the entire softmax distribution extends to various safety-critical domains, such as autonomous vehicles, medical diagnosis, and financial risk assessment, where the consequences of incorrect predictions can be severe \cite{michelmore2018evaluating, leibig2017leveraging}. By considering the full distribution of predictions, more informed and reliable decisions can be made, reducing the risk of catastrophic failures.

The Misclassification Likelihood Matrix (MLMs) proposed in this study we believe has potential to further enhance the interpretability and risk mitigation capabilities of softmax probabilities. MLMs provide a comprehensive view of the model's misclassification tendencies by capturing the likelihood of each class being misclassified as another class. By analyzing the patterns and magnitudes of these misclassification likelihoods, decision-makers can identify the most common and critical sources of error. This information can be used to prioritize model improvements, guide data collection efforts, and establish decision thresholds based on the acceptable level of risk. Moreover, MLMs can be employed to generate explanations for the model's predictions, highlighting the classes that are most likely to be confused and the factors contributing to the misclassifications. This transparency enables users to better understand the limitations and potential failure modes of the model, promoting informed decision-making and fostering trust in the system.
    

\section{Methods}


\subsection{Neural Networks}

\begin{figure}[h]
    \centering
    \includegraphics[width=0.99\columnwidth]{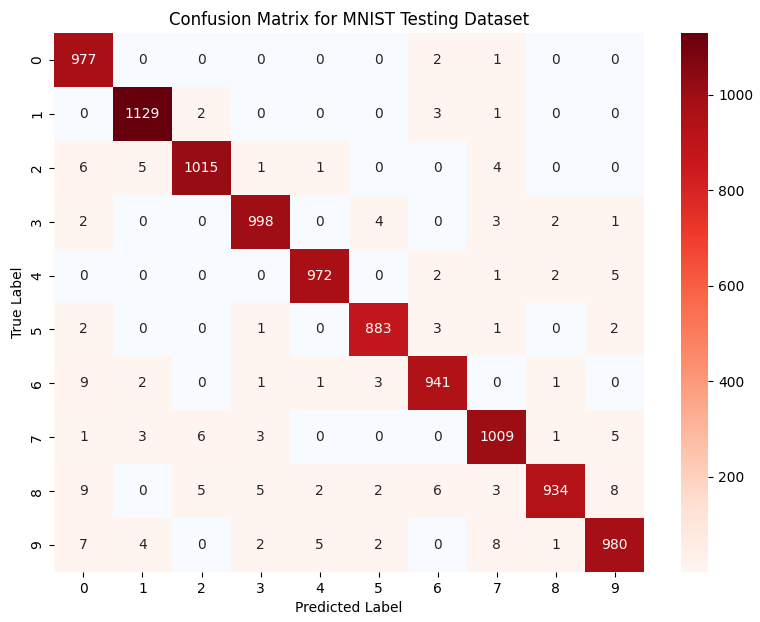}   
    \caption{MNIST testing dataset confusion matrix where the least misclassified digit is 0 (top row) and the most misclassified digit is 8 (second row from bottom up), where the figures represent a snapshot i.e. one single testing dataset, the main difference between a traditional confusion matrix and the MLM which may represent n distribution-shifted testing datasets.}
\label{fig:mnist_testing_dataset_confusion_matrix}
\end{figure}

A Convolutional Neural Network (CNN) is used to classify handwritten digits from the MNIST dataset, consisting of 60,000 training images and 10,000 testing images, each of size 28x28 grayscale (single channel) pixels, representing digits from 0 to 9.

In our analysis, we consider the classifier neural network's entire output, not just the highest probability score. The network produces a probability vector $\mathbf{p} = (p_1, p_2, \dots, p_K)$, where $\sum p_i = 1$,

This probability distribution is derived from a logit vector $\mathbf{z} = (z_1, z_2, \dots, z_K)$ through the application of the softmax function:

\begin{equation}
p_i = \text{softmax}(z_i) = \frac{e^{z_i}}{\sum_{j=1}^{K} e^{z_j}}
\end{equation}

For example, given 

$\mathbf{p} = [0.01, 0.01, 0.01, 0.01, 0.9, 0.01, 0.01, 0.01, 0.01, 0.01]$, the predicted class is '4', corresponding to the highest value at index five, reflecting the confidence of the prediction for each class from '0' to '9'.

The relationship between logits and probabilities can be expressed as:

\begin{equation}
z_i = \log \left(\frac{p_i}{1 - p_i}\right)
\end{equation}

Here, $z_i$ represents the logit for class $i$, while $p_i$ denotes the probability that the input belongs to class $i$. These logits can be interpreted as log-likelihoods of class membership \cite{goodfellow2016deep, bishop2006pattern}.

By examining the entire output vector, rather than just the highest score, we gain deeper insights into the model's decision-making process and its level of certainty or uncertainty across all possible classes.

The CNN architecture, implemented using PyTorch, consists of two convolutional layers followed by two fully connected layers. The first convolutional layer has 16 filters with a kernel size of 3x3 and a padding of 1. The second convolutional layer has 32 filters with the same kernel size and padding. Each convolutional layer is followed by a ReLU activation function and a max-pooling layer with a pool size of 2x2. The output of the second convolutional layer is flattened and passed through two fully connected layers with 128 and 10 neurons, respectively. The final output is passed through a log-softmax function to obtain the predicted class probabilities. The total number of parameters for the CNN MNIST classifier is 206,922.

The model was trained using the Stochastic Gradient Descent (SGD) optimizer with a learning rate of 0.01, batch size of 64 and Cross-Entropy loss function. 


\subsection{Perturbed MNIST Dataset}
\label{methods:PerturbedMNISTDataset}

To test our trained model, we augment by two orders of magnitude the MNIST testing dataset, where the original testing dataset consists of 10,000 images with 10 classes of approximately 1,000 examples each. We create 12 perturbations - Brightness, Contrast, Defocus Blur, Fog, Frost, Gaussian Noise, Impulse Noise, Motion Blur, Pixelation, Shot Noise, Snow and Zoom Blur - following prior work by \cite{hendrycks2019benchmarking}, at 10 intensity levels, generating a dataset of 1,210,000 images. We modify perturbation function parameters by trial and error such that increased perturbation levels on the testing dataset cause network predictive accuracies to decrease linearly. The 60,000 image training dataset was left unperturbed. While it would be desirable to use more than one dataset, it has been demonstrated that softmax output distances to centroids (explained in the next section) are greater for misclassified examples in both MNIST/CNN and CIFAR-10/ViT (Vision Transformer) \cite{sikar2024acceptautomatedpredictionsdefer}. CNNs have been the traditional go-to architecture for image recognition tasks, leveraging local receptive fields and spatial hierarchies. Vision Transformers, on the other hand, represent a more recent paradigm shift in computer vision, adapting the self-attention mechanism from natural language processing to image analysis. Despite these fundamental differences in architecture, both types of models exhibit similar behavior in terms of the relationship between softmax output distances and misclassification likelihood. 

It is also important to note that the probability distribution of a digit belonging to a class is not normally distributed, but rather follows a multinomial distribution across the 10 digit classes.


\subsection{Misclassification Likelihood Matrix}
\label{miss_class_matrix}

\begin{figure}[h]
    \centering
    \includegraphics[width=0.99\columnwidth]{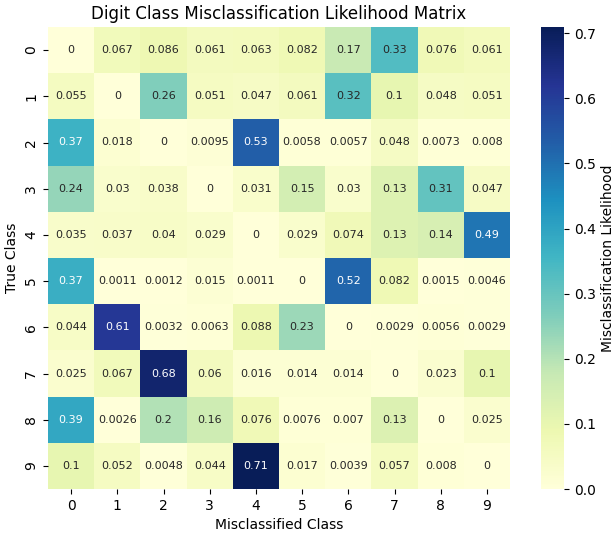}   
    \caption{Perturbed MNIST testing dataset MLM, where dark blue indicates greater, and lighter yellow lower likelihood of misclassification. The figures effectively represent 120 MNIST testing datasets with 10 levels of noise across 12 different perturbation types.} 
    \label{fig:DigitClassMisclassificationLikelihoodMatrix}
\end{figure}

To generate a Misclassification Likelihood Matrix for a particular model and dataset, we proceed as described next:

\textbf{a}. Obtaining the Centroids:
Let $ D = {(x_1, y_1), (x_2, y_2), ..., (x_n, y_n)} $ be the training dataset, where $ x_i \in \mathbb{R}^d $ represents the input features and $y_i \in {0, 1, ..., 9}$ represents the corresponding digit class label. We train a neural network classifier $f_\theta(x)$ on this dataset, where $\theta$ represents the learned parameters of the model.
To obtain the initial class centroids, we first collect the softmax outputs of the trained classifier for each correctly classified example in the training set. Let $S_c = {s_1, s_2, ..., s_m}$ be the set of softmax outputs for examples correctly classified as class $c$, where $s_i \in \mathbb{R}^{10}$. We calculate the initial centroid $\mu_c$ for class $c$ as the mean of the softmax outputs:

\begin{equation}
\mu_c = \frac{1}{m} \sum_{i=1}^m s_i
\label{eq:mean_softmax_output}
\end{equation}

We repeat this process for each digit class $c \in {0, 1, ..., 9}$ to obtain the initial centroids $\mu_0, \mu_1, ..., \mu_9$.

After obtaining the initial centroids, we use them to initialize the K-Means algorithm. The K-Means algorithm iteratively assigns each softmax output to the nearest centroid and updates the centroids based on the assigned outputs until convergence. This process refines the centroids and yields the final adjusted centroids $\hat{\mu}_0, \hat{\mu}_1, ..., \hat{\mu}_9$.

\textbf{b}. Obtaining the Distance to a Neighboring Centroid:
Given a test example $x$ with true class label $y$, we obtain the softmax output $s = f_\theta(x)$, that we interpret as points in a simplex, which has a natural Euclidean geometry. To calculate the distance between $s$ and a neighboring class centroid $\hat{\mu}c$, where $c \neq y$, we use the Euclidean distance:

\begin{equation}
d(s, \hat{\mu}c) = \sqrt{\sum\limits_{i=1}^{10} (s_i - \hat{\mu}{c,i})^2}
\label{eq:euclidean_distance_to_neighboring_centroid}
\end{equation}

This distance measures how close the softmax output of the test example is to the adjusted centroid of a different class.

\textbf{c}. Obtaining the Nearest Distances Between Digit Classes Matrix:
To construct the matrix of nearest distances between digit classes, we iterate over all test examples and calculate the nearest distance to each neighboring class centroid. Let $X_y$ be the set of test examples with true class label $y$. For each test example $x \in X_y$, we calculate the distance $d(f_\theta(x), \hat{\mu}c)$ to each neighboring class centroid $\hat{\mu}c$, where $c \neq y$.
We then find the minimum distance among all test examples in $X_y$ to each neighboring class centroid:
\begin{equation}
D{y,c} = \min{x \in X_y} d(f_\theta(x), \hat{\mu}c)
\label{eq:min_distance_to_near_centroid}
\end{equation}
The resulting matrix $D \in \mathbb{R}^{10 \times 10}$ contains the nearest distances between each pair of digit classes, where $D{y,c}$ represents the nearest distance from class $y$ to class $c$.

\textbf{d}. Obtaining the Misclassification Likelihood Matrix:
To obtain the misclassification likelihood matrix, we transform the nearest distance matrix $D$ by taking the reciprocal of each element and normalizing the rows:
\begin{equation}
L_{y,c} = \frac{\frac{1}{D_{y,c}}}{\sum\limits_{c=0}^9 \frac{1}{D_{y,c}}}
\label{eq:reciprocal_distance_matrix_eg_likelihood_matrix}
\end{equation}

The resulting matrix $L \in \mathbb{R}^{10 \times 10}$ contains the misclassification likelihoods between each pair of digit classes, where $L_{y,c}$ represents the likelihood of an example from class $y$ being misclassified as class $c$. Higher values indicate a higher likelihood of misclassification, while lower values indicate a lower likelihood.

\subsection{Testing the hypothesis}

To test the hypothesis that some misclassifications are more likely to occur than others, we can analyze how the misclassification likelihoods change as we introduce perturbations to the input images. By comparing the misclassification likelihoods at different perturbation levels, we can determine if certain digit pairs are consistently more prone to misclassification.

Let $L^{(p)} \in \mathbb{R}^{10 \times 10}$ be the misclassification likelihood matrix at perturbation level $p$, where $p \in \{1, 2, ..., 10\}$. Each element $L^{(p)}_{y,c}$ represents the likelihood of an example from class $y$ being misclassified as class $c$ at perturbation level $p$.

To obtain $L^{(p)}$, we follow a similar process as described in Section \ref{miss_class_matrix}.d, but we use the distances calculated from the perturbed examples at level $p$. Let $X^{(p)}_y$ be the set of test examples with true class label $y$ at perturbation level $p$. For each test example $x \in X^{(p)}_y$, we calculate the distance $d(f_\theta(x), \mu_c)$ to each neighboring class centroid $\mu_c$, where $c \neq y$.

We then find the minimum distance among all test examples in $X^{(p)}_y$ to each neighboring class centroid:

\begin{equation}
D^{(p)}_{y,c} = \min_{x \in X^{(p)}_y} d(f_\theta(x), \mu_c)    
\end{equation}

The resulting matrix $D^{(p)} \in \mathbb{R}^{10 \times 10}$ contains the nearest distances between each pair of digit classes at perturbation level $p$.

To obtain the misclassification likelihood matrix $L^{(p)}$, we apply the same transformation as in Section \ref{miss_class_matrix}.d:
\begin{equation}
L^{(p)}_{y,c} = \frac{\frac{1}{D^{(p)}_{y,c}}}{\sum\limits_{c=0}^9 \frac{1}{D^{(p)}_{y,c}}}    
\end{equation}

We repeat this process for each perturbation level $p$ to obtain a set of misclassification likelihood matrices $\{L^{(1)}, L^{(2)}, ..., L^{(10)}\}$.

To test the hypothesis, we can analyze the changes in misclassification likelihoods across different perturbation levels. We can calculate the average misclassification likelihood for each digit pair $(y, c)$ across all perturbation levels:
\begin{equation}
\bar{L}_{y,c} = \frac{1}{10} \sum_{p=1}^{10} L^{(p)}_{y,c}    
\end{equation}

We can then identify the digit pairs with the highest average misclassification likelihoods. If these pairs consistently have high misclassification likelihoods across different perturbation levels, it suggests that they are more prone to misclassification compared to other digit pairs.

To quantify the consistency of misclassification likelihoods across perturbation levels, we can calculate the standard deviation of the misclassification likelihoods for each digit pair:
\begin{equation}
\sigma_{y,c} = \sqrt{\frac{1}{10} \sum_{p=1}^{10} (L^{(p)}_{y,c} - \bar{L}_{y,c})^2}    
\end{equation}

A low standard deviation indicates that the misclassification likelihood for a digit pair remains relatively stable across perturbation levels, supporting the hypothesis that certain misclassifications are more likely to occur consistently.
    

\section{Results and Discussion}

\subsection{Neural Network Training}

The MNIST dataset was preprocessed using a transformation pipeline that converted the images to PyTorch tensors and normalized the pixel values to have a mean of 0.5 and a standard deviation of 0.5. The dataset was then loaded using PyTorch's DataLoader, for batch processing and shuffling of the data. The CNN MNIST classifier was trained for 10 epochs, and took 5m6s to train on a Dell Precision Tower 5810 with a 6 core Intel Xeon Processor and 32GB memory running Ubuntu 18.04. The accuracy obtained on the training dataset is 98.38\% and 98.46\% on the testing dataset.

The confusion matrix shown in Figure \ref{fig:mnist_testing_dataset_confusion_matrix} is obtained from the original testing dataset (10,000 images) i.e. without perturbations and provides some insights into what could be the most likely MNIST misclassifications. We note that number zero is the digit least, and number eight is the digit most misclassified. Other digits that have low misclassification counts are digits one and five. Digit eight has higher misclassification counts for predicted labels zero, six and nine. Digit three has relatively high misclassification counts for predicted labels five and seven.


\subsection{Clustering results}

\begin{table}[htbp]
\centering
\caption{Euclidean Distances between Class Average Softmax and K-Means Centroids}
\label{tab:diff_softmax_avg_dist_to_centroids}
\begin{tabular}{|c|c|}
\hline
Class & Distance \\
\hline
0 & 1.1694e-14 \\
\hline
1 & 1.0959e-14 \\
\hline
2 & 6.7639e-15 \\
\hline
3 & 1.7808e-14 \\
\hline
4 & 7.7425e-15 \\
\hline
5 & 1.3045e-04 \\
\hline
6 & 2.3118e-04 \\
\hline
7 & 1.3083e-14 \\
\hline
8 & 1.4139e-04 \\
\hline
9 & 1.2047e-14 \\
\hline
\end{tabular}
\end{table}

After running the K-Means clustering algorithm, initialised with the  average softmax outputs for each class, we obtain cluster centroids. We are interested in knowing how much the average softmax outputs moved from their initial position. 

Table \ref{tab:diff_softmax_avg_dist_to_centroids} presents the Euclidean distances between the average softmax outputs and the K-Means centroids for each digit class (0-9). For most digit classes (0, 1, 2, 3, 4, 7, 9), the Euclidean distances are extremely small, on the order of 1e-14 to 1e-15, indicating that the average softmax outputs are very close to the centroids obtained from the K-Means clustering algorithm. However, for digit classes 5, 6, and 8, the Euclidean distances are slightly larger, on the order of 1e-4, suggesting a minor difference between the average softmax outputs and the K-Means centroids for these specific classes, which could be attributed to the presence of some misclassified or ambiguous examples within these classes.
Despite the slight deviations observed for classes 5, 6, and 8, the overall distances between the average softmax outputs and the K-Means centroids are relatively small across all digit classes, indicating a strong similarity between the representations learned by the neural network and the centroids obtained from the K-Means clustering algorithm. The small distances suggest that using the average softmax outputs as initial centroids for the K-Means clustering algorithm is an effective approach, as they provide a good starting point for the clustering process and can potentially lead to faster convergence. Given the data is highly skewed towards the centroids, alternative statistical methods like standard percentiles lack granularity.


\subsection{Misclustered Images}


\begin{figure}[H]
    \centering
    \includegraphics[width=0.99\columnwidth]{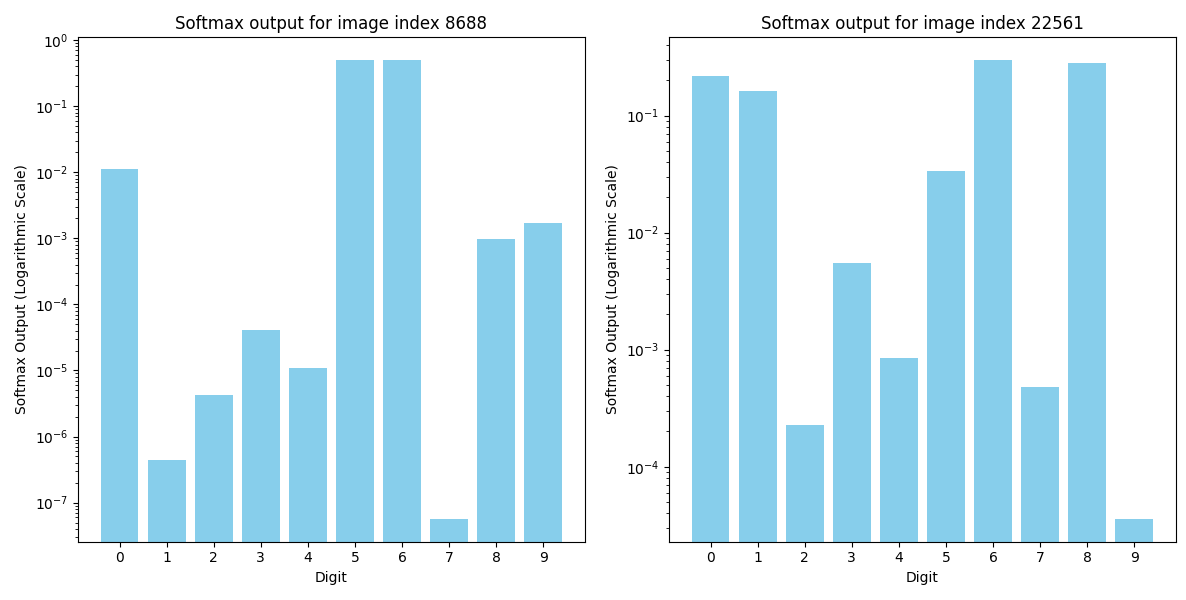}
    \caption{The Softmax output for the two "misclustered images in the correctly classified MNIST training dataset. Both are labelled as digit six. The example on the left is assigned to cluster 5 while the example on the right is assigned to cluster 8}
    \label{fig:8688_22561_Softmax_Output}
\end{figure}

Looking at the pattern of the softmax output (network prediction) 
indicates how confident a model is of the class prediction. Figure \ref{fig:8688_22561_Softmax_Output} contains two bar charts that show the MNIST-trained CNN softmax output for MNIST dataset digit ids 8688 and 22561 (starting from index 0), where both are labelled as class 6 and correctly classified by the CNN as such. The bar chart on the left shows the softmax output assigns a very high probability to both classes 5 and 6 (where 6 is the highest). The bar chart on the right shows that the CNN assigns a high probability to classes 6 and 8 (where 6 is again the highest). The cluster labels however do not correspond as digit id 8688 is nearer to the cluster 5 centroid, and digit id 22561 is nearer to cluster 8 centroid. This volatility may be explained by the relatively high deltas shown in Table \ref{tab:diff_softmax_avg_dist_to_centroids}, where the K-Means centroids moved a relatively large distance with respect to the initial values (given by average softmax outputs for correct digit classifications in the training dataset), for digit classes 5, 6 and 8. 

\begin{figure}[H]
    \centering
    \includegraphics[width=0.99\columnwidth]{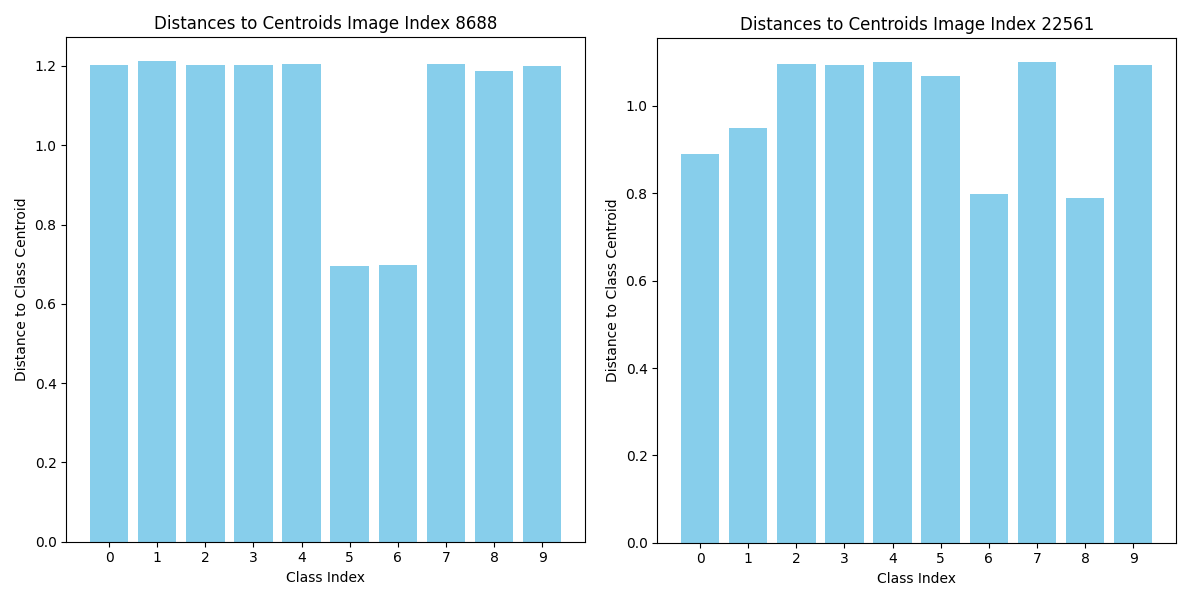}
    \caption{The Softmax prediction distance to centroids for MNIST training images indexes 8688 (left) and 22561 (right), where both images are digit class 6, and the left image is misclassified as digit class 5, while the right image is misclassified as digit class 8.}
    \label{fig:distances_to_centroids_8688_22561} 
\end{figure}

Conversely, the pattern of the distance to class centroids for a given softmax output 
indicates how close the softmax output is to a class centroid with respect to other class centroids. Figure \ref{fig:distances_to_centroids_8688_22561} shows on the left the distance to all class centroids of digit id 8688 network softmax output, placing it closing to centroids 5 and 6, while the image on the right (digit id 22561) places the softmax output closer to centroids 6 and 9. We notice the inverse relation between both sets of figures, where the highest bars (network confidence) in the first set correspond to the lowest bars (distance to class centroid) set in the second set.



\subsection{Applying Perturbations}
\begin{figure}[h] 
\begin{center}
\includegraphics[width=0.99\columnwidth]{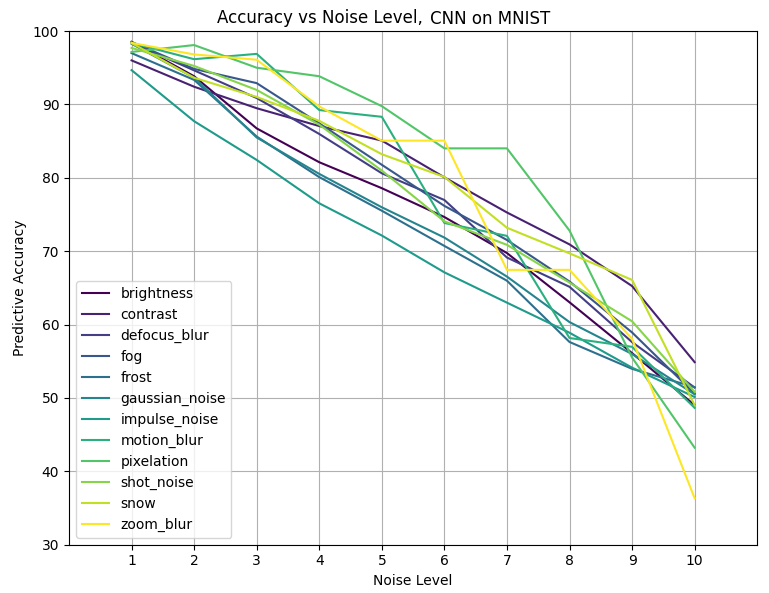}
\end{center}
\caption{Plot for network accuracy degradation for individual perturbation types: Brightness, Contrast, Defocus Blur, Fog, Frost, Gaussian Noise, Impulse Noise, Motion Blur, Pixelation, Shot Noise, Snow and Zoom Blur.}
\label{fig:Accuracy_vs_Noise_types}
\end{figure}

\begin{figure*}[h!]
    \centering
    \includegraphics[width=0.99\textwidth]{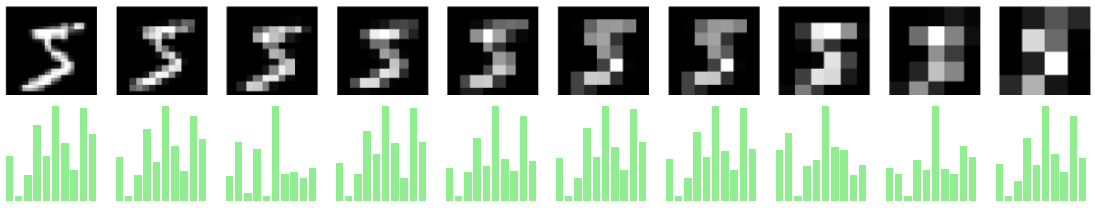}   
    \caption{Top row: MNIST class digit 5 subject 1 to 10 levels of pixelation going left to right. Bottom row: the average softmax output for the perturbed MNIST dataset class digit 5 for each perturbation level across all perturbation types. The change in bar heights is an indication of how images are moving closer and further to centroids in the cluster as noise increases.}\label{fig:Pixelation_Digit_5_images_histogramsx10_plus_softmax}
\end{figure*}

Through trial and error, we find sets of parameter values
for each perturbation function, such that as the level increases from 1 to
10, the network predictive accuracy decreases from 98.52\%
maximum to 36.32\% minimum and from average 97.61\%
(level 1) to 48.83\% (level 10). The accuracy degradation can be observed both Figures 
\ref{fig:Accuracy_vs_Noise_types} and \ref{fig:Pixelation_Digit_5_images_histogramsx10_plus_softmax}.

Figure \ref{fig:Pixelation_Digit_5_images_histogramsx10_plus_softmax} shows digit class 5 subject to increasing levels of pixelation from left to right. The bottom row of the figure shows the average softmax output for all perturbation types at a given level. We note that at all levels from 1 to 10, the highest prediction is digit class 5 (sixth column from left to right on every plot, the first being for digit 0).

\begin{figure*}[h!]
    \centering
    \includegraphics[width=0.99\textwidth]{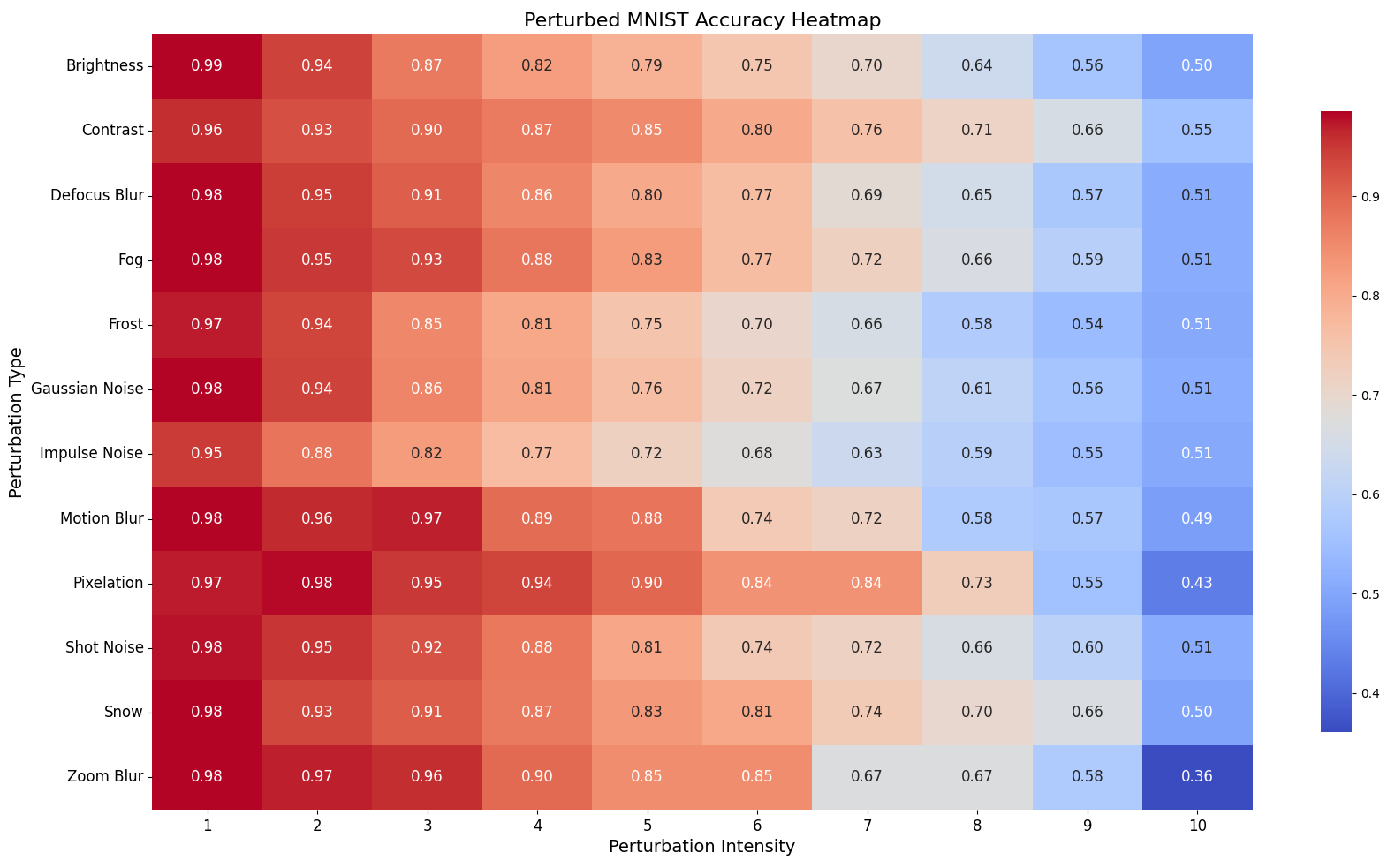}   
    \caption{Perturbed MNIST Accuracy Heatmap, where the accuracies are for all digit classes, given a perturbation type and intensity.}
    \label{fig:PerturbedMNISTAccuracyHeatmap}
\end{figure*}

Figure \ref{fig:PerturbedMNISTAccuracyHeatmap} shows a heatmap where every cell represents one perturbation (y axis labels, 12 in total) at one intensity level (x axis labels, 10 in total) applied to the 10,000 image MNIST training dataset, the perturbed images are presented to the trained model and the resulting accuracy is placed in the corresponding cell. The heatmap therefore represents an experiment with 1,210,000 predictions on perturbed images of the perturbed dataset as described in section \ref{methods:PerturbedMNISTDataset}.

The heatmap uses a color scale from red to blue, with red indicating higher accuracy values and blue representing lower accuracy. As the perturbation intensity increases, the accuracy decreases for all perturbation types.

Some perturbations, such as brightness, contrast, and defocus blur, have a more gradual impact on accuracy as the intensity increases. For example, the accuracy under brightness perturbation drops from 0.99 at intensity 1 to 0.50 at intensity 10. Other perturbations, like pixelation and zoom blur, cause a more rapid decrease in accuracy at higher intensities.

Interestingly, certain perturbations, such as snow and impulse noise, show a slight increase in accuracy at lower intensities (e.g., intensity 2) compared to the baseline (intensity 1). This suggests that a small amount of these perturbations might act as a form of data augmentation, potentially improving the model's robustness.

The model appears to be most resilient to perturbations like brightness, contrast, and fog, maintaining relatively high accuracy even at higher intensities. On the other hand, perturbations such as pixelation, shot noise, and zoom blur have a more substantial negative impact on accuracy, with values dropping below 0.6 at higher intensities.


\subsection{Misclassification Likelihood Matrix}

Figure \ref{fig:DigitClassMisclassificationLikelihoodMatrix} shows the Misclassification Likelihood Matrix obtained by the method described in section \ref{miss_class_matrix}. Comparing the figure with the MNIST testing (non-perturbed) dataset shown in Figure \ref{fig:mnist_testing_dataset_confusion_matrix} we can notice some information added by the MLM that may help inform risk assessment. On the top right of the confusion matrix is a cluster of zeros, representing digits 0, 1 and 2 misclassifications of 8 and 9. The MLM is able to provide fine-grained information about the misclassification likelihood, assigning relatively low albeit non-zero likelihood misclassification values.
As the MLM represents a much larger dataset, there are some noticeable discrepancies. The MLM does not assign a high likelihood for digit 9 misclassified as 7, which is the highest in the testing dataset. On the other hand, row 7 is a close match with 7 being misclassified as 2 holding in the perturbed MNIST dataset.

MLM and traditional confusion matrices offer complementary insights into a neural network's classification performance. The MLM, derived from multiple perturbed versions of the testing set, provides a fine-grained view of misclassification tendencies. It represents the likelihood of each digit class being misclassified as another, with color intensity indicating misclassification probability. In contrast, the confusion matrix shows actual counts of correct and incorrect classifications for a single test set.
The MLM offers an additional perspective, showing likelihood values even for rare misclassifications.

While the confusion matrix clearly shows classification performance with higher numbers on the diagonal indicating better accuracy, the MLM requires a different interpretation. Zero values on the diagonal mean one class cannot be misclassified as itself. Both tools highlight patterns like the high likelihood of misclassifying 7 as 2, while the MLM also shows likelihood for corresponding zero values in the confusion matrix e.g. the likelihood of class zero being misclassified as classes 1, 2, 3, 4 or 5.




\section{Conclusions and Future Work}

In this study, we proposed a novel approach for quantifying the reliability of neural network predictions under distribution shifts by leveraging clustering techniques and analyzing the distances between softmax outputs and class centroids. 

The introduction of the Misclassification Likelihood Matrix (MLM) provides a comprehensive view of the model's misclassification tendencies, enabling decision-makers to identify the most common and critical sources of errors.
The results highlight the potential of MLMs in enhancing the interpretability and risk mitigation capabilities of softmax probabilities.

The implications of our work extend beyond image classification, with ongoing applications in autonomous systems, such as self-driving cars, to improve the safety and reliability of decision-making in complex, real-world dynamic environments. 
By identifying scenarios where human judgment is preferable to the autonomous system's assessment, our methodology aims to mitigate the risks associated with distribution shifts and enhance the overall trustworthiness of automated decision-making systems.

Future directions: 

1. Incorporate entropy and entropy apex metrics into our framework. The entropy apex, a point approximately equidistant to all class centroids, is expected to be a region of high entropy where misclassified examples cluster. 

2. Explore alternative distance metrics, such as cosine similarity, to measure the relationships between class centroids and predictions. This could offer new perspectives on the geometric properties of the feature space and how they relate to classification decisions.

3. Integrate our metric with complementary techniques, including uncertainty estimation and domain adaptation methods. This integration could lead to a more comprehensive framework for handling distribution shifts, combining the strengths of multiple approaches to create more robust and adaptable models.

4. Investigate the application of our methodology to a broader range of real-world autonomous systems, focusing on how it can enhance safety and reliability in diverse and challenging environments.    

\printbibliography                

\end{document}